\documentclass[letterpaper]{article} 
\usepackage{natbib}
\setcitestyle{numbers,square}

\usepackage{aaai2026}
\usepackage{times}  
\usepackage{helvet}  
\usepackage{courier}  
\usepackage[hyphens]{url}  
\usepackage{graphicx} 
\def\UrlFont{\rm}  
\usepackage{natbib}  
\usepackage{caption} 
\usepackage{algorithm}
\usepackage{algorithmic}
\usepackage{amsthm}
\usepackage{amsmath}
\usepackage{booktabs}
\usepackage[T1]{fontenc}
\usepackage{amssymb}
\usepackage{tikz}
\usepackage{multirow}
\usetikzlibrary{positioning,arrows.meta}

\usepackage{xcolor} 

\newcommand{\mixupx}{\tilde{x}}

\newtheorem{theorem}{Theorem}
\newtheorem{proposition}{Proposition}
\usepackage{newfloat}
\usepackage{listings}
\DeclareCaptionStyle{ruled}{labelfont=normalfont,labelsep=colon,strut=off} 
\floatstyle{ruled}
\newfloat{listing}{tb}{lst}{}
\floatname{listing}{Listing}
\nocopyright
\title{SelectMix: Enhancing Label Noise Robustness through Targeted Sample Mixing}

\author {
    Qiuhao Liu\textsuperscript{\rm 1},
    Ling Li\textsuperscript{\rm 1},
    Yao Lu\textsuperscript{\rm 2},
    Qi Xuan\textsuperscript{\rm 2},
    Zhaowei Zhu\textsuperscript{\rm 2,}$^3$,
    Jiaheng Wei \textsuperscript{\rm 1}\thanks{Corresponding author: jiahengwei@hkust-gz.edu.cn}
}
\affiliations {
    \textsuperscript{\rm 1}The Hong Kong University of Science and Technology (Guangzhou)\\
    \textsuperscript{\rm 2}BIAI, ZJUT \qquad $^3$D5Data.ai\\
    
}

\usepackage{bibentry}

\begin{document}

\maketitle

\begin{abstract}
Deep neural networks tend to memorize noisy labels, severely degrading their generalization performance. Although Mixup has demonstrated effectiveness in improving generalization and robustness, existing Mixup-based methods typically perform indiscriminate mixing without principled guidance on sample selection and mixing strategy, inadvertently propagating noisy supervision. To overcome these limitations, we propose \textit{SelectMix}, a confidence-guided mixing framework explicitly tailored for noisy labels. \textit{SelectMix} first identifies potentially noisy or ambiguous samples through confidence-based mismatch analysis using $K$-fold cross-validation, then selectively blends identified uncertain samples with confidently predicted peers from their potential classes. Furthermore, \textit{SelectMix} employs soft labels derived from all classes involved in the mixing process, ensuring the labels accurately represent the composition of the mixed samples, thus aligning supervision signals closely with the actual mixed inputs. Through extensive theoretical analysis and empirical evaluations on multiple synthetic (MNIST, Fashion-MNIST, CIFAR-10, CIFAR-100) and real-world benchmark datasets (CIFAR-N, MNIST and Clothing1M), we demonstrate that \textit{SelectMix} consistently outperforms strong baseline methods, validating its effectiveness and robustness in learning with noisy labels.
\end{abstract}

\section{Introduction}
Large‑scale vision datasets scraped from the web or crowd-sourcing often suffer from imperfect labels~\cite{deng2009imagenet,xiao2015learning,zhang2016understanding, wei2021learning, xiao2017fashion,liu2023humans, wei2023aggregate,liu2024automatic}. When training with noisy labels, overparameterized nets tend to first fit the clean subset and then memorize the wrong annotations, so training loss keeps falling while test accuracy stalls~\cite{liu2020early,xia2020robust}. This growing gap between empirical and true risk threatens various applications such as autonomous driving, medical diagnosis, and content moderation~\cite{goldberger2017training}, requiring learning algorithms that exploit abundant data, yet remain resistant to noisy labels. 

Existing research on noise‑robust training has converged on two primary directions. \emph{Loss‑centric methods} modify the objective to reduce the impacts of potentially mislabeled samples~\cite{zhang2018generalized,wang2019symmetric, lin2024learning,bai2021understanding}, while \emph{Sample‑centric methods} instead decide which data to trust or treat noise as a semi‑supervised problem, selecting or relabeling samples on the fly~\cite{han2018co,li2020dividemix,wei2020combating,cordeiro2023longremix}. A complementary strand leverages strong data augmentation, i.e., Mixup~\cite{zhang2017mixup} and Manifold Mixup~\cite{verma2019manifold}, which interpolate inputs or hidden representations and their labels, flattening decision boundaries and delaying memorization.

Although Mixup is an effective and efficient augmentation strategy on clean datasets, it can inevitably harm model performance when the training labels suffer from a relatively large noise rate. For example, on CIFAR‑10 with $40$\% symmetric label noise, vanilla Mixup reduces top‑$1$ test accuracy by $12$ percentage points compared to standard empirical risk minimization~\cite{liu2023over}. A similar decline appears for CutMix and Manifold Mixup when noisy and clean samples are interpolated indiscriminately~\cite{hwang2024rc}. Additional empirical studies~\cite{liu2023over,hwang2024rc} also reveal that when noisy and clean samples are mixed indiscriminately, the erroneous supervision can be propagated rather than diluted, leading to accuracy degradation in high-noise-rate scenarios. These drops occur because incorrect labels are mixed up, so error signals spread instead of being diluted. Recent attempts to model label uncertainty or to tune mixing ratios lessen the damage but still choose partners at random, leaving the core vulnerability untouched. Hence, our aim is to deal with the following tough task when Mixup meets noisy labels: 
\textbf{What samples should be selected and how to mix them up?}

Briefly speaking, we propose \emph{SelectMix}: \emph{Select}ing potentially mislabeled samples and \emph{Mix}ing them up with samples belonging to the possible class, ensuring that all classes covered in the mixed soft labels appear in the mixed sample.
Specifically, for every training sample, we perform lightweight $K$‑fold inference and flag those whose predicted label disagrees with the given annotation as noise candidates, a strategy inspired by repeated cross‑validation detectors~\cite{chen2024detecting} and the principled noise‑estimation philosophy of Confident Learning~\cite{northcutt2021confident}. Rather than discarding such instances, \textit{SelectMix} linearly interpolates each ``mismatched'' sample with a peer sharing its predicted class with high‑confidence, and assigns a soft target that balances noisy and predicted labels. This pairing preserves Mixup’s boundary‑smoothing effect while preventing the cross‑class error propagation that arises when partners are chosen randomly.
The key contributions of this work are:
\begin{itemize}
    \item \textbf{}We introduce \textit{SelectMix}, a novel approach that selectively applies Mixup-based data augmentation in the presence of noisy labels. By identifying samples that are likely to have incorrect labels through confidence-based filtering and mismatch detection, \textit{SelectMix} strategically mixes these samples with others from potentially confusing categories, enhancing robustness and generalization in noisy label scenarios.
    \item We provide theoretical analysis illustraing that \textit{SelectMix} eliminates the class-dependent bias term and shrinks the instance-dependent variance term in the Mixup risk decomposition. The result offers the first formal guarantee that mismatch-guided partner choice yields measurable robustness to label noise. 
    \item Empirical validation of \textit{SelectMix} on MNIST, Fashion‑MNIST, CIFAR-10, CIFAR-100, CIFAR‑N, and Clothing1M demonstrates its effectiveness in handling label noise across various datasets. 
\end{itemize}

\section{Related Work}
There are two lines of related work that are most relevant to us. The first traces the evolution of Mixup-based data augmentation, from simple pixel blends to sophisticated label interpolation. The second surveys noise-robust learning, covering objectives that down-weight corrupted labels and strategies that identify or relabel a trustworthy subset. 
 
\subsection{Mixup-Based Data Augmentation}
Mixup~\cite{zhang2017mixup} is a highly successful technique to improve the generalization of neural networks by augmenting the training data with combinations of random pairs. Following this seminal work, numerous variants have emerged~\cite{yun2019cutmix,sun2022lumix,wang2022adamix,zhao2023mixpro,baek2021gridmix,noh2023rankmixup,hong2021stylemix,tu2022guidedmix,islam2024diffusemix,yang2022recursivemix,wei2022smooth}. As classified in a recent survey~\cite{jin2024survey}, previous research in Mixup method can be broadly categorized into two groups: \textit{Sample Mixup Policies} and \textit{Label Mixup Policies}, more details come as follows.

\paragraph{Sample Mixup Policies} 
    Sample Mixup techniques can be viewed along a spectrum of \emph{what} is blended and \emph{how} the blending policy is chosen. Early work applies a fixed linear interpolation to raw images; AdaMixup \cite{wang2022adamix} relaxes this static rule by adaptively learning the mixing coefficient during training, enabling the generated samples to avoid manifold intrusion. Moving from pixels to features, Manifold Mixup \cite{verma2019manifold} interpolates hidden representations in latent space, enriching internal decision boundaries without altering the input domain. A second family performs spatial composition. CutMix \cite{yun2019cutmix} pastes a rectangular patch from one image onto another, GridMix \cite{baek2021gridmix} replaces the rectangle with a regular grid, and ResizeMix \cite{qin2020resizemix} first downsizes the source image so the pasted object remains intact. This line differs from Cutout \cite{devries2017improved}, which simply removes a patch and therefore discards class information. Patch selection can be guided by saliency. PuzzleMix \cite{kim2020puzzle} uses saliency maps and an optimal-transport solver to rearrange informative regions, while StyleMix \cite{hong2021stylemix} goes a step further by disentangling content and style features, mixing them independently to boost diversity. The idea extends naturally to semi-supervised learning. GuideMix \cite{tu2022guidedmix} interpolates labeled and unlabeled images, letting mutual information flow from the labeled side and thus improving pseudo-mask quality. Most recently, DiffuseMix \cite{islam2024diffusemix} combines natural images with diffusion-generated counterparts that share the same structural cues, producing realistic yet label-consistent augmentations. Taken together, these methods demonstrate that increasingly sophisticated policies, ranging from learnable ratios and latent-space mixing to spatial cuts, saliency guidance, and generative synthesis, all fall under the shared principle of ``mix to augment.''

\paragraph{Label Mixup Policies}
Label Mixup Policies focus on refining the labels of mixed samples to ensure consistency during training. Optimizing calibration methods, such as DivideMix~\cite{li2020dividemix}, mitigate label noise by treating likely noisy samples as unlabeled data, thus enhancing generalization. RankMixup~\cite{noh2023rankmixup} introduces a ranking-aware regularization that preserves the confidence hierarchy between raw and mixed samples. TokenMix~\cite{liu2022tokenmix} and TokenMixup~\cite{choi2022tokenmixup} used the raw samples attention scores for calucating the $\lambda$ of classes. MixPro~\cite{zhao2023mixpro} argued that scores obtained by the model in the early stages are inaccurate and give incorrect information thus proposed to combine with the region and attention scores to calculate the $\lambda$. DivideMix~\cite{li2020dividemix} drops the labels of the samples that are most likely to be noisy and uses the noisy samples as unlabeled data to regularize the model, avoiding overfitting and improving the generalization performance. LUMix~\cite{sun2022lumix} addresses label noise in Mixup by adaptively adjusting the mixing ratio using prediction-based confidence and injecting uniform noise to simulate label uncertainty.

\begin{figure*}[!htb]
    \centering
    \includegraphics[width=\textwidth]{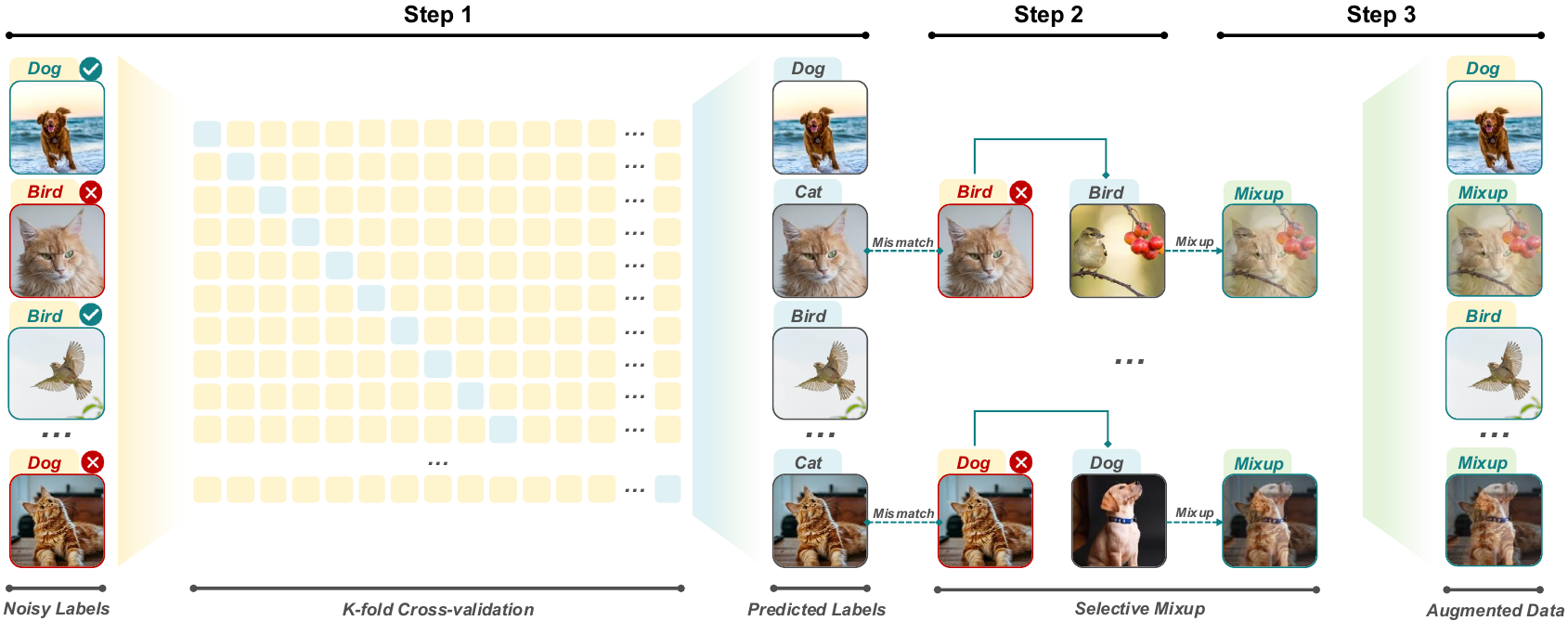}
    \caption{Pipeline of \textit{SelectMix}: \textbf{Step 1) Predict.} $K$-fold cross‑validation yields an out‑of‑fold label for every image. \textbf{Step 2) Select.} Flag samples with $\text{pred} \neq \text{noisy}$ as \textit{mismatches}; the rest are \textit{reliable}. \textbf{Step 3) Mix.} For each mismatch, pick a reliable image from its predicted class and apply Mixup; the target is $\lambda y^{\text{noisy}} + (1-\lambda)y^{\text{pred}}$.}
    \label{fig:selectmix}
  \end{figure*}

\subsection{Learning with Noisy Labels}
Existing research on noisy label learning largely falls into two complementary strands. 
\paragraph{Loss-Correction Methods} These techniques redesign the training objective so that mislabeled examples exert less influence. Early work reweights the sampling distribution to down‑scale high‑loss instances~\cite{liu2015classification}. Subsequent studies propose noise‑robust surrogates for cross‑entropy~\cite{wei2020optimizing,wei2023fairness}, such as unhinged and sigmoid losses~\cite{natarajan2013learning,ghosh2017robust}, Generalized Cross Entropy (GCE)~\cite{zhang2018generalized}, MAE‑based hybrid objectives~\cite{amid2019robust}, and Symmetric Cross Entropy (SCE)~\cite{wang2019symmetric}. A related line estimates a class‑transition matrix and explicitly corrects the loss~\cite{xia2019anchor,hendrycks2018using, li2022estimating,kye2022learning,zhang2021learning}. Although these objectives are model‑agnostic and computationally light, their effectiveness hinges on accurate transition estimates or carefully tuned hyper‑parameters, both of which are hard to obtain under instance‑dependent corruption~\cite{cheng2020learning,zhu2021second}.

\paragraph{Sample-Selection and Semi-Supervised Methods} An alternative view treats noisy labels as missing or uncertain and seeks a clean subset for supervision~\cite{zhu2021clusterability,zhu2022detecting}. Co-Teaching~\cite{han2018co} trains two networks that exchange low-loss samples, leveraging the observation that deep models memorise clean data before noisy annotations. DivideMix~\cite{li2020dividemix} builds on this idea: it fits a Gaussian mixture to per-sample losses, splits data into clean and noisy partitions, and uses MixMatch~\cite{berthelot2019mixmatch} to refine pseudo-labels, yielding state-of-the-art results. More recent variants incorporate uncertainty calibration~\cite{karim2022unicon} or long-tail rebalancing~\cite{cordeiro2023longremix}.

Despite their success, the interaction between data augmentation and noise-robust learning remains under-explored. In particular, it is unclear \emph{when} and \emph{why} \textit{Mixup} improves robustness in noisy settings. Our work addresses this question by analysing Mixup’s failure modes under label noise and introducing a mismatch-guided partner selection strategy that preserves its boundary-smoothing benefits without propagating erroneous supervision. We hope these findings will inspire a deeper theoretical understanding of Mixup and inform its principled use in future noisy label research.



\section{Method}

In this section, we introduce \textit{SelectMix} when learning with noisy labels, 
which requires no noise‑rate estimate, adds negligible computational overhead, and adapts automatically to varying noise levels.

\subsection{SelectMix}
An overview of the method is shown in Figure \ref{fig:selectmix}. The detailed steps come as follows.



\paragraph{Cross-Validated Prediction and Mismatch Identification}

We first train a base model $g_1$ using $K$-fold cross-validation on the noisy training set $\mathcal{D} = \{(x_i, y_i^{\text{noisy}})\}_{i=1}^N$, where $N$ denotes the number of samples. Let $C$ denote the number of classes. For each sample $x_{i}$ we record its out-of-fold soft prediction
$\hat{\mathbf{p}}_{i}=g_{1}^{\text{oof}}(x_{i})\in[0,1]^{C}$
and set the surrogate clean label:
\begin{equation}
    y_{i}^{\text{pred}}
=\arg\max_{c\in\{1,\dots,C\}}\hat{\mathbf{p}}_{i,c}.
\end{equation}
We then define the mismatch index set:
\begin{equation}
    \mathcal{M} = \{i \mid y_i^{\text{noisy}} \neq y_i^{\text{pred}}\},
\end{equation}
which likely contains mislabeled samples. To facilitate targeted Mixup, we construct a class-wise clean sample index:
\begin{equation}
    \mathcal{I}[c] = \{j \mid y_j^{\text{pred}} = c\},
\end{equation}
which maps each class $c$ to a set of samples confidently predicted as class $c$.

\paragraph{SelectMix Augmentation}
During training, each mini-batch sample $(x_i, y_i^{\text{noisy}}, y_i^{\text{pred}})$ is processed differently depending on its membership in $\mathcal{M}$. For samples without label disagreement, we use the original input and noisy label directly. For samples in $\mathcal{M}$, we first retrieve a reference sample $x_j$ from the clean index $\mathcal{I}[y_i^{\text{noisy}}]$, ensuring label consistency. We then draw a mixing coefficient $\lambda$ from a Beta distribution and construct a mixed input:
\begin{equation}
    \tilde{x}_i=\lambda x_i+(1-\lambda)x_j,
\end{equation}
\begin{equation}
    \tilde{y}_i=\lambda y_i^{\text{noisy}}+(1-\lambda)y_i^{\text{pred}}.
\end{equation}
Importantly, we do not treat the label as a hard target. Instead, we assign a soft label that interpolates between the noisy and predicted labels.

This interpolation produces a soft label $\tilde{y}_i$ balancing the information in the original annotation and the
model-based surrogate. The network should respect this balance in proportion to the mixing weight $\lambda$; we therefore minimise a Mixup-style composite loss.
\begin{theorem}\label{thm:soft-ce}
Let $\tilde y=\alpha y^{(1)}+(1-\alpha)y^{(2)}$ be a convex
combination of two one-hot vectors and let $f_\theta(\tilde x)$ be a
softmax prediction. Then
$$
\ell\!\bigl(f_\theta(\tilde x),\tilde y\bigr)
=\alpha\,\ell\!\bigl(f_\theta(\tilde x),y^{(1)}\bigr)
+(1-\alpha)\,\ell\!\bigl(f_\theta(\tilde x),y^{(2)}\bigr).\vspace{-2pt}
$$
\end{theorem}
The proof follows directly from the linearity of cross-entropy in its
second argument. Applying
Theorem \ref{thm:soft-ce} with
$\alpha=\lambda$, $y^{(1)}=y_i^{\text{noisy}}$ and
$y^{(2)}=y_i^{\text{pred}}$ gives the practical form we have:
\begin{equation}
\mathcal{L} = \lambda \cdot \ell(f_\theta(\tilde{x}), y_i^{\text{noisy}}) + (1 - \lambda) \cdot \ell(f_\theta(\tilde{x}), y_i^{\text{pred}}),
\end{equation}
where $\ell(\cdot, \cdot)$ denotes the standard cross-entropy loss.

Notably, by interpolating not only the inputs but also the supervision targets, \textit{SelectMix} reduces the risk of overfitting to noisy annotations and guides the network to favor patterns supported by both the data and its learned prediction structure. Algorithm 1 in Appendix A gives the full pseudocode for
\noindent\textit{SelectMix}.

\section{Theoretical Analysis}
\label{sec:theory-full}

This section demonstrates that \emph{SelectMix} reduces the population cross-entropy risk compared with vanilla Mixup considering the realistic \textbf{instance-dependent noise (IDN)} setting. In the IDN setting, the noisy label $\tilde y$ is produced by an instance-specific flipping rule, that is, $\Pr(\tilde y\mid y,x)$ may vary with the input $x$.  For contrast, \textbf{class-dependent noise} (CDN) assumes a class-level transition that depends only on the clean label $y$ and is independent of the particular instance.  The analysis below keeps this distinction explicit when decomposing the risk. 

\subsection{Notation and Noise Model}
Denote by $x\in\mathcal X$ an input and its clean one--hot label by $ y \in \{e_1,\dots,e_C\}$.  The observed noisy label is $\tilde y=y+\check y$ where the \emph{noise residue} satisfies $\sum_k \check y_k=0$.  We only assume the standard IDN property $\mathbb{E}[\check y\mid x]=m(x)$ with global mean $\mathbb{E}[m(x)]=\mathbf 0$; no parametric form for $m(\cdot)$ is required.

Note that $f_\theta:\mathcal X\to\Delta^{N}$ is a softmax network and $\ell(f_\theta(x),\tilde y)= -\tilde y^{\top}\ln f_\theta(x)$ indicates the cross--entropy loss. For two i.i.d.~samples we mix inputs and labels with $\lambda\sim\text{Beta}(\alpha,\alpha)$:
\begin{equation}
x_{\text{mix}} = \lambda x + (1-\lambda) x',
\label{eq:mix}
\end{equation}
\begin{equation}
    \quad
\tilde y_{\text{mix}} = \lambda \tilde y + (1-\lambda) \tilde y'.
\end{equation}

\begin{proposition}[Risk decomposition of Mixup under IDN]
    The resulting population risk is $R_{\text{mix}}=\mathbb{E}[\ell(\lambda,f_\theta(x_{\text{mix}}),\tilde y_{\text{mix}})]$. We can rewrite the population risk of Mixup in 
\begin{equation}
  R_{\text{mix}}
  \;=\;
  R_{\text{clean}}
  \;+\;
  \underbrace{\kappa_{\text{IDN}}\,R_{\text{IDN}}}_{\text{IDN term}}
  \;+\;
  \underbrace{\kappa_{\text{CDN}}\,R_{\text{CDN}}}_{\text{CDN term}},
  \label{eq:mixloss-split-new}
\end{equation}

with coefficients
\[
  \kappa_{\text{IDN}}
  =\mathbb{E}_{\lambda}\!\bigl[\lambda^{2} + (1-\lambda)^{2}\bigr],
  \kappa_{\text{CDN}}
  =2\,\mathbb{E}_{\lambda}\!\bigl[\lambda(1-\lambda)\bigr].
\]
$R_{\text{clean}}$, $R_{\text{IDN}}$ and $R_{\text{CDN}}$ denote the loss on clean mixed labels, the population risk of IDN and the population risk of CDN, respectively. 

Please refer to Appendix B for detailed proof.


\end{proposition}

\subsection{SelectMix Formulation}
Note that a \textbf{mismatch} is detected when the model prediction disagrees with the observed label:
\begin{equation}
\mathcal M(x)=\mathbf 1\{\arg\max f_\theta(x) \neq \arg\max \tilde y\}.
\end{equation}
Reliable points satisfy $\mathcal M(x)=0$.  For such $x$ we keep the sample intact ($\lambda=1$).  For a mismatch ($\mathcal M(x)=1$) we draw a partner $x_r$ from the reliable pool that shares the same predicted class and mix the pair with the usual $\lambda$.  The partner label is taken as clean ($y_r=y$). Let $\rho=\Pr(\mathcal M=1)$ denote the global mismatch rate. With the help of Appendix C, the population risk of \textit{SelectMix} can be written as:
\begin{equation}
  R_{\text{sel}}
  = R_{\text{clean}}
  + \rho\,\kappa_{\text{IDN}}\,R_{\text{IDN}},
  \label{eq:sel-risk-new}
\end{equation}
where the CDN term vanishes and the IDN term is reduced by the mismatch
rate.
\subsection{Risk Gap Under Weak Reliability}
Assume the network’s average log-likelihood on the reliable set
exceeds random guessing by at least a margin \(\delta>0\): 
\begin{equation}
\mathbb{E}\big[y^{\top}\ln f_\theta(x) \mid \mathcal M=0\big] \ge \mathbb{E}\big[\tfrac1C\mathbf 1^{\top}\ln f_\theta(x)\big] + \delta.
\label{eq:delta-new}
\end{equation}
This ``weak reliability'' holds after a brief warm-up in practice.  Using Eqn.\eqref{eq:mixloss-split-new} and Eqn.\eqref{eq:sel-risk-new}
together with Eqn.\eqref{eq:delta-new} (combined with Pinsker’s inequality to lower-bound $R_{\text{CDN}}$) for any $0<\lambda<1$, we obtain the central inequality
\begin{equation}
  R_{\text{sel}}
  \;\le\;
  R_{\text{mix}}
  -\kappa_{\text{CDN}}\;\delta\,\rho.
  \label{eq:gap-new}
\end{equation}
Since $\kappa_{\text{CDN}}$ is maximised when the $\text{Beta}$ prior is
peaked at $\lambda=0.5$, \textit{SelectMix} benefits most from uniform mixing and
from higher mismatch rate $\rho$.  When $\rho=0$ Ineqn.
\eqref{eq:gap-new} becomes an equality, showing that \textit{SelectMix} never
hurts on clean data.

\paragraph{Interpretation.}
Eqn. \eqref{eq:gap-new} formalises the intuition that selecting
partners by prediction–label mismatch removes cross-class bias and
limits instance-specific noise, while preserving Mixup’s boundary
smoothing.

\section{Experiments}


In this section, we extensively validate our method on five benchmark datasets, namely CIFAR-10N, CIFAR-100N~\cite{wei2021learning}, MNIST~\cite{lecun1998mnist}, Fashion-MNIST~\cite{xiao2017fashion} and Clothing1M~\cite{xiao2015learning}. 

\subsection{Datasets and Implementation Details}


\begin{table}[!t]
\centering
\resizebox{\linewidth}{!}{
\begin{tabular}{l c|cccc|c}
\toprule
\textbf{Method} & & \multicolumn{4}{c|}{\textbf{Sym}} & \textbf{Asym} \\
\cmidrule{3-6}
               & & 20\% & 40\% & 60\% & 80\% & 40\% \\
\midrule
No Mixup & Best & 86.11 & 82.99 & 74.89 & 31.83 & 87.33 \\
      & Last & 82.27 & 64.94 & 42.22 & 17.54 & 80.49 \\
\midrule
Mixup & Best & 91.63 & 84.37 & 80.22 & 56.89 & 89.31 \\
      & Last  & 87.04 & 69.53 & 48.03 & 21.55 & 81.91  \\
\midrule
Mixup* & Best & 92.17 & 90.01 & 84.71 & 40.79 & 90.96 \\
        & Last & 91.92 & 89.92 & 83.97 & 38.39 & 90.68 \\
\midrule
StyleMix& Best & 91.32 & 88.20 & 82.58 & 56.75 & 81.54 \\
        & Last & 86.19 & 71.89 & 56.31 & 37.73 & 79.71 \\
\midrule
Cutmix & Best & 92.36 & 90.06 & 83.24 & 48.66 & 90.48 \\
        & Last & 87.04 & 71.93 & 50.31 & 34.57 & 73.62 \\
\midrule
RecursiveMix & Best & 93.49 & 90.49 & 84.62 & 47.40 & 91.30 \\
        & Last & 85.05 & 68.12 & 44.95 & 44.54 & 80.68 \\
\midrule
PuzzleMix & Best & 86.47 & 90.30 & 84.13 & 47.67 & 90.98 \\
        & Last & 80.44 & 81.82 & 68.55 & 17.42 & 81.82 \\
\midrule
ManifoldMix & Best & 93.16 & 88.62 & 82.90 & 47.30 & 91.04 \\
        & Last & 87.48  & 67.99 & 47.25 & 43.06 & 81.18 \\
\midrule
GuideMix & Best& \textbf{93.60} & 89.94 & 84.31 & \textbf{56.95} & 89.87 \\
        & Last & 84.30 & 65.80 & 53.71 & \textbf{44.73} & 79.77 \\
\midrule
Cutout& Best & 86.35 & 88.56 & 83.44 & 48.27 & 83.92 \\
        & Last & 81.17 & 66.32  & 53.34 & 16.74 & 81.89 \\
\midrule
DiffuseMix & Best & 85.88 & 77.29 & 70.03 & 41.67 & 81.22 \\
           & Last & 75.12 & 55.18 & 33.77 & 16.98 & 67.09 \\
\midrule
\midrule
\textbf{SelectMix}&Best & 92.64 & \textbf{91.27} & \textbf{86.23} & 44.43 & \textbf{91.53} \\
 & Last & \textbf{92.50} & \textbf{90.96}  &  \textbf{85.25} & 42.11  & \textbf{91.07 }\\
\bottomrule
\end{tabular}
}
\caption{Comparison with state-of-the-art methods in test accuracy on CIFAR-10 with symmetric and asymmetric noise. Last row shows final 10-epoch average test accuracy.}
\label{tab:CIFAR synthetic noise}
\end{table}

We adopt four types of real-world noise for CIFAR-10N: Rand1, Rand2, Rand3, and Worst, with corresponding noise rates of 17.302\%, 18.146\%, 17.644\%, and 40.456\%, respectively. For CIFAR-100N, we use one type of noise, Noisy-Fine, with a noise rate of 40.216\%. Following previous work~\cite{li2020dividemix}, we also experiment two type of synthetic noise namely: \textit{symmetric} and \textit{asymmetric}. Symmetric noise is introduced by uniformly flipping the labels to any of the other classes with a probability $p$. In contrast, asymmetric noise is designed to resemble real-world scenarios, where labels are only flipped to semantically similar classes (\textit{e.g.} bird$\leftrightarrow$airplane, dog$\leftrightarrow$cat).

Following~\cite{bai2021understanding}, we employ a ResNet‑18 for CIFAR‑10, Fashion‑MNIST, and MNIST, and a ResNet‑34 for CIFAR‑100. We train the network using SGD with a momentum of 0.9, weight decay of 0.0001, and a batch size of 128. All models are trained for 200 epochs with an initial learning rate of 0.1, which is reduced by a factor of 10 at the $100$th and $150$th epochs. For Clothing1M we adopt the ResNet‑$18$ backbone and training protocol of~\cite{wei2020combating}. Models are trained for $15$ epochs with a batch size of $64$. The learning rate is scheduled as follows: $8 \times 10^{-4}$ for epochs $1-5$, $5 \times 10^{-4}$ for epoch $6-10$ and $5 \times 10^{-5}$  for epochs $11-15$.

\begin{table}[!t]
\centering
\resizebox{\linewidth}{!}{
\begin{tabular}{l c|cccc|c}
\toprule
\textbf{Method} & & \multicolumn{4}{c|}{\textbf{CIFAR-10N}} & \textbf{100N} \\
               & & Rand1& Rand2& Rand3 & Worst& Noisy100\\
\midrule
No Mixup& Best & 86.14 & 85.93 & 86.05 & 77.24 & 46.30 \\
        & Last & 82.45 & 81.60 & 82.95 & 62.45 & 42.12 \\
\midrule
Mixup & Best & 91.11 & 90.80 & 90.99 & 82.44 & 51.42 \\
        & Last & 87.36 & 86.47 & 87.87 & 68.25 & 50.16 \\
\midrule
Mixup* & Best & 91.96 & 92.39 & 91.95 & 88.33 & 60.39 \\
        & Last & 91.68 & 92.19 & 91.71 & 87.98 & 59.75 \\
\midrule
StyleMix& Best & 91.30 & 90.85 & 90.80 & 86.87 & 57.34 \\
        & Last & 87.22 & 85.54 & 86.52 & 72.79 & 49.74 \\
        
\midrule
Cutmix& Best & 92.65 & 92.32 & 92.59 & 87.72 & 58.49 \\
        & Last & 88.41 & 87.24 & 87.35 & 72.95 & 51.22 \\
\midrule
RecursiveMix& Best & \textbf{93.56} & \textbf{93.33} & 93.45 & 88.57 & 61.49 \\
        & Last & 85.57 & 85.04 & 85.50 & 65.71 & 51.70 \\
\midrule
PuzzleMix& Best & 93.30 & 93.30 & \textbf{93.56} & 88.81 & \textbf{64.73} \\
        & Last & 89.67 & 89.62 & 90.03 & 76.64 & 57.07 \\
\midrule
ManifoldMix& Best & 92.80 & 91.47 & 92.31 & 87.36 & 58.28 \\
        & Last & 85.97 & 86.01 & 86.77 & 65.55 & 53.87 \\
\midrule
GuideMix & Best& 92.76 & 92.56 & 92.69 & 87.75 & 61.97 \\
        & Last & 86.95 & 86.39 & 87.16 & 67.14 & 53.56 \\
\midrule
Cutout& Best & 93.20 & 92.82 & 92.52 & 88.74 & 52.04 \\
        & Last & 82.98 & 83.09 & 83.15 & 63.86 & 48.26 \\
\midrule
DiffuseMix& Best & 84.77 & 83.26 & 84.36 & 74.92 & 43.02 \\
        & Last & 75.04 & 74.69 & 77.42 & 54.19 & 38.24 \\
\midrule
\midrule
\textbf{SelectMix}&Best & 93.26 & 93.16 & 93.41 & \textbf{90.48} & 64.04 \\
 & Last & \textbf{93.13} & \textbf{93.09} & \textbf{93.27} & \textbf{90.21} & \textbf{62.25} \\
\bottomrule
\end{tabular}
}
\caption{Performance comparison on CIFAR-10N and CIFAR-100N. Last row shows final 10-epoch average accuracy (\%).}
\label{tab:comparison cifarn}
\end{table}

\begin{table*}[!t]
\centering
\setlength{\tabcolsep}{4pt}  
\begin{tabular}{l c|cccc|cccc}
\toprule
\multirow{4}*{\textbf{Method}}
&
& \multicolumn{4}{c|}{\textbf{MNIST}}
& \multicolumn{4}{c}{\textbf{Fashion-MNIST}} \\ \cmidrule{3-10}
& &
  \multicolumn{3}{c|}{\textbf{Sym}} & \multicolumn{1}{c|}{\textbf{Asym}} &
  \multicolumn{3}{c|}{\textbf{Sym}} & \multicolumn{1}{c}{\textbf{Asym}} \\ \cmidrule{3-10}
& & 20\% & 50\% & 80\% & 40\% & 20\% & 50\% & 80\% & 40\% \\
\midrule
No Mixup      & Best & 99.61 & 99.37 & 98.24 & 99.49 & 93.31 & 91.74 & 84.61 & 92.63 \\
              & Last & 99.58 & 99.27 & 97.64 & 99.29 & 93.22 & 91.46 & 82.48 & 92.28 \\
\midrule
Mixup         & Best & 99.65 & 99.48 & 96.75 & 99.55 & 94.41 & 92.20 & 85.31 & 93.47 \\
              & Last & 97.81 & 71.17 & 32.68 & 97.68 & 91.68 & 72.26 & 53.02 & 89.10 \\
\midrule
Mixup*        & Best & 99.71 & 99.46 & 98.66 & 99.52 & 94.00 & 92.70 & 86.25 & 93.24 \\
              & Last & 99.65 & 99.39 & 97.91 & 99.34 & 93.61 & 92.51 & 84.22 & 92.97 \\
\midrule
StyleMix      & Best & 99.47 & 99.35 & 98.34 & 99.08 & 93.51 & 92.25 & 85.54 & 92.34 \\
              & Last & 59.83 & 72.24 & 49.49 & 59.35 & 84.49 & 73.72 & 59.41 & 83.04 \\
\midrule
CutMix        & Best & 99.58 & 99.46 & 98.27 & 99.42 & 94.86 & 93.69 & 85.74 & 94.24 \\
              & Last & 99.17 & 96.80 & 93.90 & 97.12 & 93.17 & 86.59 & 81.41 & 86.25 \\

\midrule
RecursiveMix  & Best & 99.71 & 99.61 & 98.21 & \textbf{99.72} & 94.98 & \textbf{93.85} & 86.68 & 94.24 \\
              & Last & 97.04 & 86.82 & 77.10 & 94.67 & 89.91 & 73.84 & 76.15 & 84.95 \\
\midrule
PuzzleMix     & Best & 24.07 & 28.90 & 36.34 & 25.91 & 39.63 & 39.49 & 37.83 & 35.89 \\
              & Last & 13.15 & 14.94 & 16.10 & 16.06 & 23.26 & 28.73 & 20.80 & 22.39 \\
\midrule
ManifoldMix   & Best & 99.66 & 99.47 & 98.04 & 99.59 & 94.78 & 92.94 & 84.20 & 93.61 \\
              & Last & 98.40 & 80.34 & 48.57 & 97.04 & 90.00 & 69.56 & 62.22 & 85.29 \\

\midrule
GuideMix      & Best & 99.65 & 99.44 & 98.24 & 99.58 & 94.89 & 93.30 & 85.38 & 94.39 \\
              & Last & 97.96 & 74.51 & 46.03 & 95.13 & 88.92 & 67.46 & 57.36 & 84.89 \\

\midrule
Cutout        & Best & 99.70 & 99.60 & 98.79 & 99.67 & 94.44 & 93.33 & 85.95 & 93.62 \\
              & Last & 99.06 & 98.37 & 94.84 & 98.90 & 88.01 & 85.81 & 67.16 & 66.18 \\
\midrule
DiffuseMix    & Best & 98.92 & 97.86 & 90.44 & 97.17 & 88.73 & 84.12 & 70.47 & 84.56 \\
              & Last & 78.43 & 63.21 & 22.69 & 82.91 & 77.65 & 67.89 & 48.12 & 72.35 \\
\midrule
\midrule
\textbf{SelectMix} & Best & \textbf{99.73} & \textbf{99.62} & \textbf{98.88} & 99.55 
                   & \textbf{95.02} & 93.72 & \textbf{86.73} & \textbf{94.85} \\
                   & Last & \textbf{99.70} & \textbf{99.56} & \textbf{98.66} & \textbf{99.46}
                   & \textbf{94.69} & \textbf{93.43} & \textbf{85.62} & \textbf{93.95} \\
\bottomrule
\end{tabular}
\caption{Test accuracy (\%) under symmetric and asymmetric noise on MNIST (left block) and Fashion-MNIST (right block). Last row shows final 10-epoch average accuracy(\%).}
\label{tab:mnist_fmnist_combined}
\end{table*}

\subsection{Comparison with Existing Mixup-based Methods}

We compare \textit{SelectMix} with multiple Mixup baselines~\cite{zhang2017mixup, hong2021stylemix, yun2019cutmix, yang2022recursivemix, kim2020puzzle, verma2019manifold, tu2022guidedmix, devries2017improved,islam2024diffusemix} using the same network architecture under the same experirmental settings described above. We denote the experiments as \textit{Mixup, Mixup*\footnote{\textbf{Mixup*} is obtained in two steps: (i) perform a single $K$-fold cross-validation on the noisy training set and record the out-of-fold predicions  $\{y_i^{\text{pred}}\}_{i=1}^N$ from the base model $g_1$; (ii) apply the standard Mixup procedure, but replace each noisy label $y^{\text{noisy}}_i$with its prediction $y^{\text{pred}}_i$. }, StyleMix, Cutmix, RecursiveMix, PuzzleMix, ManifoldMix, GuideMix, Cutout, DiffuseMix} in the experiment tables.

\paragraph{Experiment Results on CIFAR-10}Table \ref{tab:CIFAR synthetic noise} shows the results on CIFAR-10 dataset and CIFAR-100 dataset with different types and level of synthetic label noise ranging from 20\% to 80\%. We report the best test accuracy for all epochs. \textit{SelectMix} outperforms state-of-the art methods under moderate noise ratio while performing mediocre in both low and high noise ratio. Specifically, on the CIFAR-10 dataset \textit{SelectMix} achieves the highest peak accuracy at the challenging 40\% and 60\% symmetric noise levels and at 40\% asymmetric noise, while recording the best last-epoch accuracy under every setting tested (20\%, 40\%, 60\% symmetric and 40\% asymmetric), highlighting both its strength and stability. This highlights \textit{SelectMix}’s robustness to increasing noise severity.  Under asymmetric noise, which more closely resembles real-world scenarios, \textit{SelectMix} attains 91.53\% accuracy, surpassing all baselines. Notably, even under extreme symmetric noise (80\%), where most methods experience significant degradation, \textit{SelectMix} maintains competitive performance (44.43\%), suggesting enhanced noise resistance in challenging conditions.

\paragraph{Experiment Results on CIFAR-10N and CIFAR-100N} As shown in Table \ref{tab:comparison cifarn}, on the CIFAR-N dataset with real-world noisy labels, our proposed \textit{SelectMix} demonstrates strong and consistent performance in all noisy settings. On CIFAR-10N, it achieves the best average last-epoch accuracies surpassing all competing methods. Notably, while \textit{RecursiveMix} and \textit{PuzzleMix} slightly outperform \textit{SelectMix} in some of the best epoch metrics, their performance drops significantly in the last-epoch average, indicating potential overfitting or instability in later training stages. In contrast, \textit{SelectMix} maintains a narrow performance gap between its best and last epoch, confirming its resilience under noisy supervision. Additionally, on CIFAR-100N, \textit{SelectMix} further expands the lead, achieving a 2.50\% improvement in last-epoch accuracy compared to the best-performing baseline, while maintaining competitive best-epoch results. These findings highlight \textit{SelectMix}’s scalability and its effectiveness in real-world noisy label learning scenarios.

\paragraph{Experiment Results on MNIST}Table \ref{tab:mnist_fmnist_combined} (left block) compares \textit{SelectMix} with ten strong augmentation baselines under four noise settings. Under moderate noise (20\% symmetric) \textit{SelectMix} attains 99.73\% best accuracy. Although the margin is small, the gap in last‑epoch accuracy is larger: 99.70\% versus 99.06\%, indicating better training stability. The advantage widens as the corruption intensifies. With 50\% symmetric noise, \textit{SelectMix} improves the best accuracy from 99.61\% (RecursiveMix) to 99.62\%, and raises the last‑epoch figure by 12.74 percentage points over the strongest baseline (Cutmix, 86.82\%). At 80\% symmetric noise—the hardest setting—most methods collapse; StyleMix, StyleCutMix and ManifoldMix lose more than 45 percentage points between best and last checkpoints. \textit{SelectMix}, in contrast, still delivers 98.88\%/98.66\% best/last accuracy, surpassing the next best last‑epoch score (Cutmix, 93.90\%) by 4.76 percentage points. Across all noise regimes, \textit{SelectMix} maintains a narrow best‑epoch lead and, more importantly, preserves that lead to the end of training, confirming that mismatch‑guided partner selection mitigates late‑stage memorization and delivers state‑of‑the‑art robustness on MNIST.

\paragraph{Experiment Results on Fashion-MNIST} The same trend appears in Table \ref{tab:mnist_fmnist_combined} (right block). With 20\% symmetric noise, \textit{SelectMix} achieves the highest peak accuracy (95.02\%) and retains almost all of it at the end of training (94.69\%), a drop of only 0.33 percentage points compared with 1.69 percentage points for CutMix. At 50\% noise RecursiveMix briefly leads, but collapses by 20 percentage points \textit{SelectMix} secures the highest final accuracy (93.43\%), a 6.8 percentage points advantage over CutMix and 0.92 percent points over Mixup*. Under the hardest 80\% symmetric setting, Style‑based mixing variants lose more than 30 percentage points, whereas \textit{SelectMix} still ends at 85.62\%, 1.40 percentage points ahead of the next best competitor. Finally, in the 40\% asymmetric case it tops both best and last figures, finishing 0.70 percentage points ahead of RecursiveMix. Across all settings, the minimal gap between the best and last checkpoints confirms the superior robustness of \textit{SelectMix}.

\paragraph{Experiment Results on Clothing1M} Table \ref{tab:comparison Clothing1M} presents the highest precision of 1 on the real world Clothing1M benchmark, which contains approximately 38\% naturally corrupted labels. \textit{SelectMix} achieves the highest peak accuracy (69.03\%) and, more importantly, the highest final accuracy (68.72\%) among all eleven methods. A closer look reveals two key observations. First, \textit{SelectMix} enjoys a margin of +0.40 percent points over the strongest baseline at its peak (RecursiveMix, 68.63\%) and a larger +0.79 percent points at convergence over the most stable competitor (ManifoldMix, 67.93\%). Second, the method maintains its advantage throughout training: the best‑to‑last drop is only 0.31 percent points, matching the smallest decay in the table, whereas classic Mixup loses 3.31 percent points and Cutout declines by 3.32 percent points. These results indicate that mismatch‑guided partner selection curbs late‑stage memorisation and preserves performance on large‑scale, real‑noise data. \textit{SelectMix} not only secures the highest accuracy but also shows the strongest stability on Clothing1M, confirming its effectiveness under realistic label noise.
\begin{table}[!t]
    \centering
    \begin{tabular}{lcc}
        \toprule
        \textbf{Method} & \textbf{Best} & \textbf{Last} \\
        \midrule
        No Mixup       & 62.08 & 58.55 \\
        Mixup          & 67.53 & 64.22 \\
        Mixup* & 68.58 & 67.15 \\
        StyleMix       & 67.82 & 65.47 \\
        Cutmix         & 68.16 & 67.04 \\
        RecursiveMix   & 68.63 & 67.31 \\
        PuzzleMix      & 60.04 & 58.61 \\
        ManifoldMix    & 68.24 & 67.93 \\
        GuideMix       & 68.29 & 65.02 \\
        Cutout         & 68.71 & 65.39 \\
        DiffuseMix     & 59.37 & 52.03 \\
        \midrule 
        \textbf{SelectMix}      & \textbf{69.03} & \textbf{68.72} \\
        \bottomrule
    \end{tabular}
    \caption{Comparison with state-of-the-art methods in test accuracy (\%) on Clothing1M.}
    \label{tab:comparison Clothing1M}
\end{table}

\begin{table}[!t]
\centering
\begin{tabular}{l c|cccc}
\toprule
\textbf{Method} & & \multicolumn{4}{c}{\textbf{CIFAR-10N}}  \\
               & & Rand1  & Rand2 & Rand3 & Worst  \\

\midrule
\textbf{$\alpha=0.1$} & Best & 92.09 & 92.30 & 92.30 & 87.86  \\
        & Last & 89.47 & 89.98 & 90.10 & 78.14 \\
\midrule
\textbf{$\alpha=0.5$} & Best & 92.75 & 92.59 & 92.75 & 88.87  \\
        & Last & 92.47 & 92.48 & 92.53 & 88.54  \\
\midrule
\textbf{$\alpha=1.0$} & Best & \textbf{93.26} & \textbf{93.16} & \textbf{93.41} & \textbf{90.48}  \\
        & Last & \textbf{93.13} & \textbf{93.09} & \textbf{93.27} & \textbf{90.21}  \\
\midrule
\textbf{$\alpha=2.0$} & Best & 92.66 & 92.70 & 92.49 & 89.73  \\
        & Last & 92.56 & 92.63 & 92.40 & 89.30\\
\midrule
\textbf{$\alpha=4.0$} & Best & 92.62 & 92.23 & 92.77 & 88.79\\
        & Last & 92.52  & 92.10 & 92.59 & 88.57  \\
\bottomrule
\end{tabular}
\caption{Ablation study results in terms of test accuracy (\%) on CIFAR-10N.}
\label{tab:ablation study}
\end{table}

\subsection{Ablation Study}
\paragraph{Ablation on $\alpha$} We study the effect of removing different components to provide insights into what makes \textit{SelectMix} successful. We analyze the results in Table \ref{tab:ablation study} as follows.Varying the Beta parameter shows a clear sweet‑spot: $\alpha=1.0$ consistently delivers the highest peak and final accuracy across all CIFAR‑10N splits, with almost no best‑to‑last decay. More extreme settings either under‑mix ($\alpha=1.0$) and suffer late‑stage collapse, or over‑smooth ($\alpha>2$) and forfeit a full percentage point of performance, confirming that moderate, uniform mixing best balances boundary smoothing and noise dilution.

\section{Conclusion}
When Mixup meets noisy labels, we investigate what samples should be selected for mixup and how they should be mixed up. Specifically, we introduce SelectMix for robust learning with noisy labels by leveraging mismatch‑aware Mixup. SelectMix trains a single network equipped with a lightweight $K$‑fold mismatch detector and achieves noise resilience through selective sample pairing, confidence‑weighted label blending, and soft target supervision. Extensive experiments across five benchmarks demonstrate that SelectMix consistently surpasses state‑of‑the‑art mixup-based methods in addressing label noise, attaining the strongest final accuracy on Clothing1M and narrowing the best‑to‑last‑epoch gap on CIFAR‑N and other synthetic‑noise tasks. 
\bibliography{aaai2026}

\clearpage

\clearpage
\appendix
\section{Appendix A: Omitted Proofs}\label{appendix: proof of risk decomposition}
\subsection{Proof of Risk Decomposition}
We derive the risk decomposition of Mixup under instance-dependent noise.
Let $\ell(\lambda,f_\theta(x_{\text{mix}}),\tilde y_{\text{mix}})$ be the Mixup cross-entropy defined in the main text and $\phi_x=\ln f_\theta(x)$ and $\phi_{x'}=\ln f_\theta(x')$ for brevity. Expanding $\tilde y = y + \check y$ taking expectations give

$$
\begin{aligned}
&\mathbb{E}_{\lambda,\widetilde{\mathcal D}}
  \bigl[\ell\!\bigl(\lambda,f_\theta(x_{\text{mix}}),\tilde y_{\text{mix}}\bigr)\bigr] \\[2pt]
=&\;\mathbb{E}_{\lambda,\widetilde{\mathcal D}}
     \Bigl[(\lambda\tilde y+(1-\lambda)\tilde y')^{\!\top}
           \bigl(\lambda\phi_x+(1-\lambda)\phi_{x'}\bigr)\Bigr] \\[2pt]
=&\;\mathbb{E}_{\lambda,\mathcal D}
     \Bigl[(\lambda y+(1-\lambda)y')^{\!\top}
           \bigl(\lambda\phi_x+(1-\lambda)\phi_{x'}\bigr)\Bigr] \\[-2pt]
&\quad
 +\mathbb{E}_{\lambda,(x,\check y),(x',\check y')}
     \Bigl[(\lambda\check y+(1-\lambda)\check y')^{\!\top}
           \bigl(\lambda\phi_x+(1-\lambda)\phi_{x'}\bigr)\Bigr] \\[2pt]
=&\;\mathbb{E}_{\lambda,\mathcal D}
     \Bigl[(\lambda y+(1-\lambda)y')^{\!\top}
           \bigl(\lambda\phi_x+(1-\lambda)\phi_{x'}\bigr)\Bigr] \\[-2pt]
&\quad
 +\mathbb{E}_{\lambda,(x,\check y)}
    \!\bigl[(\lambda^{2}+(1-\lambda)^{2})\,\check y^{\!\top}\phi_x\bigr] \\[-2pt]
&\quad
 +2\,\mathbb{E}_{\lambda}\,
     \mathbb{E}_{x}\!\bigl[\lambda(1-\lambda)\,\mathbb{E}[\check y]^{\!\top}\phi_x\bigr] \\[2pt]
=&\;\underbrace{\mathbb{E}_{\lambda,\mathcal D}
     \Bigl[(\lambda y+(1-\lambda)y')^{\!\top}
           \bigl(\lambda\phi_x+(1-\lambda)\phi_{x'}\bigr)\Bigr]}_{\text{Clean Mixup Loss}} \\[-2pt]
&\quad
 +\underbrace{\mathbb{E}_{\lambda}\!\bigl[\lambda^{2}+(1-\lambda)^{2}\bigr]\,
     \operatorname{tr}\bigl(\operatorname{Cov}(\check y,\phi_x)\bigr)}_{\text{IDN}} \\[-2pt]
&\quad
 +\underbrace{\mathbb{E}[\check y]^{\!\top}
     \mathbb{E}_{x}[\phi_x]}_{\text{CDN}}.\\
=&R_{\text{clean}}+R_{\text{IDN}}+R_{\text{CDN}}
\end{aligned}
$$

Thus the Mixup loss separates into three disjoint contributions:
\emph{Clean Mixup Loss}, an \emph{IDN} term that depends on the covariance between noise residue and the log-softmax, and a \emph{CDN} term driven by the global label bias.  In the main text we analyse how each part behaves under \textit{SelectMix}.

\subsection{Risk of \textit{SelectMix}}\label{appendix: risk of select mix}
Because each mismatch sample $(\mathcal M(x)=1)$ is paired with a \emph{high-confidence} partner $x_r$—that is, a point for which the network’s prediction coincides with the observed label—we only assume the partner’s residue is \emph{much smaller in expectation} than that of an arbitrary noisy label:

\begin{equation}
    \bigl\lVert\mathbb E[\check y_r]\bigr\rVert_1 \le \varepsilon,
\qquad 
\text{with }\varepsilon \ll 1 .\tag{A1}
\end{equation}

Using this relaxed condition, the mixed target becomes

\begin{equation}
\begin{aligned}
  \tilde y_{\text{mix}}
    &= \lambda\bigl(y+\check y\bigr)
       + (1-\lambda)\bigl(y_r+\check y_r\bigr) \\[2pt]
    &= y + \lambda\check y + (1-\lambda)\check y_r .
\end{aligned}\notag
\end{equation}

Taking the global expectation and applying IDN $(\mathbb E[\check y]=\mathbf 0)$ together with (A1) yields

\begin{equation}
    \bigl\lVert\mathbb E[\check y_{\text{mix}}]\bigr\rVert_1
  = (1-\lambda)\,\bigl\lVert\mathbb E[\check y_r]\bigr\rVert_1
  \;\le\; (1-\lambda)\,\varepsilon .\notag
\end{equation}

Note that for a symmetric Beta prior
\(\kappa_{\text{IDN}}+\kappa_{\text{CDN}}=1\), so the two terms still
partition the total noise contribution.

Hence the class-dependent term is bounded by

\begin{equation}
    R_{\text{CDN}}^{\text{sel}}
  =\mathbb E[\check y_{\text{mix}}]^{\!\top}
     \mathbb E_{x_{\text{mix}}}[\ln f_\theta(x_{\text{mix}})]
  = \mathcal O\!\bigl(\varepsilon\bigr),\notag
\end{equation}
which is negligible when $\varepsilon$ is small (for example, after a warm-up epoch the network’s high-confidence set typically has $\varepsilon \approx 0$).  Thus SelectMix effectively nullifies the class-dependent bias even without assuming the partner is perfectly clean; it suffices that the partner be \emph{slightly more reliable} than the original noisy label.

Only the primary residue in a mismatch contributes to IDN, hence
\begin{equation}
  R_{\text{IDN}}^{\text{sel}}
  =\rho\,\kappa_{\text{IDN}}\,R_{\text{IDN}}.\notag
\end{equation}
The clean part is unchanged, so
\begin{equation}
  R_{\text{sel}}
  = R_{\text{clean}}
  + \rho\,\kappa_{\text{IDN}}\,R_{\text{IDN}}.\notag
\end{equation}
Hence, under the mild condition $\lVert\mathbb E[\check y_r]\rVert_1\!\le\!\varepsilon$ and the mismatch rate $\rho$, \textit{SelectMix} retains the clean-data benefit of Mixup, completely suppresses the class-dependent bias, and attenuates the instance-dependent term by a factor of $\rho$.

\section{Appendix B: Detail steps of \textit{SelectMix}}\label{appendix: pseudo code for selectmix}

{
\begin{algorithm}[!h]
\caption{\textit{SelectMix}. Line 1-3: Predict and mapping; Line 7-14: Mixup mismatch samples.}
\label{alg:selectivemixup}
\textbf{Input}: Training dataset $\mathcal{D} = \{(x_i, y_i^{\text{noisy}})\}_{i=1}^N$\\
\textbf{Parameter}: Mixing coefficient $\alpha$\\
\textbf{Output}: Trained model parameters $\theta$\\
\begin{algorithmic}[1]
\STATE Train model $g_1$ with $K$-fold cross-validation to get predicted labels $\{y_i^{\text{pred}}\}_{i=1}^N$
\STATE Identify mismatched samples: $\mathcal{M} = \{i \mid y_i^{\text{noisy}} \neq y_i^{\text{pred}}\}$
\STATE Build clean index map: $\mathcal{I}[c] = \{j \mid y_j^{\text{pred}} = c\}$
\STATE Initialize model parameters $\theta$ and optimizer
\FOR{each epoch $e = 1$ to $E$}
    \FOR{each mini-batch $\{(x_i, y_i^{\text{noisy}}, y_i^{\text{pred}}, i)\}$}
        \IF{$i \in \mathcal{M}$}
            \STATE Sample index $j$ from $\mathcal{I}[y_i^{\text{noisy}}]$
            \STATE Retrieve clean sample $x_j$
            \STATE Sample $\lambda \sim \text{Beta}(\alpha, \alpha)$
            \STATE $\mixupx_i \leftarrow \lambda x_i + (1 - \lambda) x_j$
        \ELSE
            \STATE $\mixupx_i \leftarrow x_i$
        \ENDIF
        \STATE Compute output: $\hat{y} = f(\mixupx_i; \theta)$
        \STATE Compute loss: $\mathcal{L} = \lambda \cdot \ell(\hat{y}_i, y_i^{\text{noisy}}) + (1 - \lambda) \cdot \ell(\hat{y}_i, y_i^{\text{pred}})$
        \STATE Backpropagate and update $\theta$ using optimizer
    \ENDFOR
\ENDFOR
\STATE \textbf{return} $\theta$
\end{algorithmic}
\end{algorithm}
}

\section{Appendix C: Experiment Details.}
This appendix summarises the data, noise generation, and training hyper-parameters used throughout our experiments.

\paragraph{Training Settings of the Used Datasets} 
Following \cite{bai2021understanding}, we train a ResNet-18 on CIFAR-10, CIFAR-10N, Fashion-MNIST, and MNIST, and a ResNet-34 on CIFAR-100 and CIFAR-100N.  All networks are optimised with SGD (momentum 0.9, weight-decay $1\times10^{-4}$, batch 128) for 200 epochs; the learning rate starts at 0.1 and is decayed by a factor of 10 at epochs 100 and 150. For Clothing1M we adopt the ResNet-18 schedule of \cite{wei2020combating}: batch 64, 15 epochs, and a three-stage learning rate of $8\times10^{-4}$ (epochs 1–5), $5\times10^{-4}$ (epochs 6–10), and $5\times10^{-5}$ (epochs 11–15).

\paragraph{Generating Noise Labels on CIFAR, MNIST and Fashion-MNIST Datasets} We adopt a symmetric noise model that generates noisy labels by randomly flipping the clean label to the other possible classes with probability $\epsilon$. And we set $\epsilon=0.2,\;0.4,\;0.6,\;0.8$ for CIFAR-10, $\epsilon=0.4$ for CIFAR-100. In contrast, asymmetric noise is designed to resemble real-world scenarios, where labels are only flipped to semantically similar classes (\textit{e.g.} bird$\leftrightarrow$airplane, dog$\leftrightarrow$cat). For MNIST and Fashion-MNIST we follow \cite{wei2020combating}, adopting symmetric noise rates of $\epsilon = 0.2,\;0.5,\;0.8$ and an asymmetric rate of $\epsilon = 0.4$.

\end{document}